\documentclass{article}

\usepackage{spconf,amsmath,graphicx}
\usepackage{color}
\usepackage{pifont}
\usepackage{soul}
\usepackage{xcolor}
\setstcolor{red}

\begin{document}

\title{Surrogate Supervision for Medical Image Analysis:\\ Effective Deep Learning From Limited Quantities of Labeled Data}
\name{\parbox{\linewidth}{\centering 
Nima Tajbakhsh$^{1}$, 
Yufei Hu$^{1}$, 
Junli Cao$^{1}$, 
Xingjian Yan$^{1}$, 
Yi Xiao$^{2}$, 
Yong Lu\sthanks{corresponding author}$^{3}$, 
Jianming Liang$^{1}$, \\
Demetri Terzopoulos$^{1}$, 
Xiaowei Ding$^{1}$}}
\address{$^{1}$ VoxelCloud, Inc. \\ $^{2}$ Department of Radiology, Changzheng Hospital, Second Military Medical University\\ $^{3}$ Radiology of Department, Ruijin Hospital, Shanghai Jiaotong University School of Medicine}

\maketitle

\begin{abstract}
We investigate the effectiveness of a simple solution to the common problem of deep learning in medical image analysis with limited quantities of labeled training data. The underlying idea is to assign artificial labels to abundantly available unlabeled medical images and, through a process known as surrogate supervision, pre-train a deep neural network model for the target medical image analysis task lacking sufficient labeled training data. In particular, we employ 3 surrogate supervision schemes, namely rotation, reconstruction, and colorization, in 4 different medical imaging applications representing classification and segmentation for both 2D and 3D medical images. 3 key findings emerge from our research: 1) pre-training with surrogate supervision is effective for small training sets; 2) deep models trained from initial weights pre-trained through surrogate supervision outperform the same models when trained from scratch, suggesting that pre-training with surrogate supervision should be considered prior to training any deep 3D models; 3) pre-training models in the medical domain with surrogate supervision is more effective than transfer learning from an unrelated domain (e.g., natural images), indicating the practical value of abundant unlabeled medical image data.
\end{abstract}
\section{Introduction}
\label{sec:intro}

The limited training sample size problem in medical imaging has often been mitigated through transfer learning, where models pre-trained on ImageNet are fine-tuned for target medical image analysis tasks. Despite some promising results~\cite{tajbakhsh16,hoo16}, this approach has its limitations. First, it may limit the designer to architectures that have been pre-trained on ImageNet, which are often needlessly deep for medical imaging, thus retarding training and inference. Second, as shown in~\cite{bau17}, fine-tuning a pre-trained model even with a large labeled dataset results in a model where a large fraction of the neurons remain ``loyal'' to the unrelated source dataset. As such, fine-tuning may not fully leverage the full capacity of a pre-trained model. Third, transfer learning is barely applicable to 3D medical image analysis applications, because the 2D and 3D kernels are not shape  compatible. Therefore, transfer learning from natural images is only a partial solution to the common problem of insufficient labeled data in medical imaging.

The limitations associated with transfer learning from a foreign domain, together with the abundance of unlabeled data in medical imaging leads to the following question: \emph{Can we pre-train network weights directly in the medical imaging domain by assigning artificial labels to unlabeled medical data?} This question has recently been addressed in mainstream computer vision, leading in a number of surrogate supervision schemes including color prediction~\cite{larsson17}, rotation prediction~\cite{gidaris18}, and noise prediction~\cite{bojanowski17}, where the key idea is to assign to unlabeled data artificial labels for the surrogate task and then use the resultant supervision, known as surrogate supervision, to pre-train a deep model for the vision task of interest. Surrogate supervision schemes have improved markedly; however, the resulting learned representation is not yet as effective as the representations learned through strong supervision. Nevertheless, surrogate supervision can still be a viable solution to the limited sample size problem in medical imaging, where strong supervision is expensive and difficult to obtain, sometimes even for small-scale datasets.

In this paper, we assess the effectiveness of surrogate supervision in tackling the limited sample size problem in medical imaging, as an alternative to training from scratch or transfer learning. Specifically, our research attempts to address the following central question: \emph{Does learning through surrogate supervision provide more effective weight initialization than random initialization or initial weights transferred from an unrelated domain?} To answer this question, we consider four medical imaging applications: false positive reduction for nodule detection in chest CT images, lung lobe segmentation in these images, diabetic retinopathy classification in fundus images, and skin segmentation in color tele-medicine images. For each application, we train models using various fractions of the training data with initial weights pre-trained using surrogate supervision, pre-trained weights transferred from ImageNet (where possible), and random weights. Our experiments on 3D datasets show that surrogate supervision leads to improved performance over training from scratch. Our experiments on 2D datasets further show a performance gain over both training from scratch and transfer learning when the medical image dataset differs markedly from natural images.

\section{Related Work}

Self-supervised learning with surrogate supervision is a relatively new trend in computer vision, with promising schemes appearing only in recent years. Consequently, the literature on the effectiveness of surrogate supervision in medical imaging is meager. Jamaludin et al.~\cite{jamaludin17} proposed longitudinal relationships between medical images as the surrogate task to pre-train model weights. To generate surrogate supervision, they assign a label of 1 if two longitudinal studies belong to the same patient and 0 otherwise. Alex et al.~\cite{varghese17} used noise removal in small image patches as the surrogate task, wherein the surrogate supervision was created by mapping the patches with user-injected noise to the original clean image patches. Ross et al.~\cite{ross18} used image colorization as the surrogate task, wherein color colonoscopy images are converted to gray-scale and then recovered using a conditional Generative Adversarial Network (GAN).

In view of the limited exploration of surrogate supervision in the medical imaging literature, it remains important to investigate this approach further, hopefully to heighten the community's awareness of its potential in overcoming the limited labeled training data problem and train better-performing medical image analysis models.

\section{Surrogate Supervision Schemes}

\begin{figure}
\includegraphics[width=0.99\linewidth]{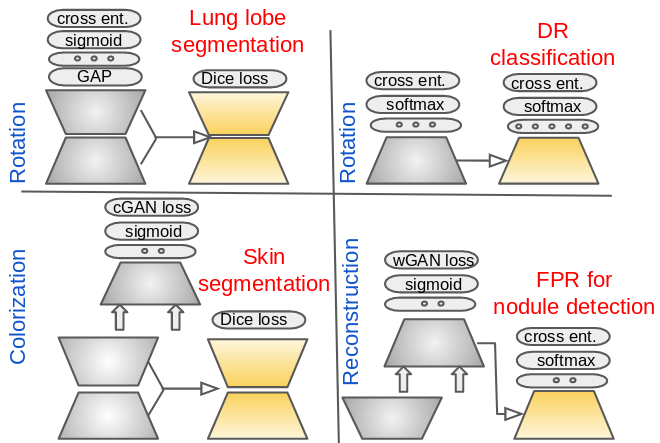}
\caption{Surrogate supervision schemes and the targets tasks used in our study. In each cell, the left network belongs to the surrogate task trained with surrogate supervision and the right network belongs to the target task. The grey and yellow trapezoids indicate the untrained weights and pre-trained weights, respectively. See Section~\ref{sec:apps} for further details.}
\label{fig:ss}
\end{figure}

We use rotation~\cite{gidaris18} as the surrogate supervision where possible. This is because of its simplicity and superior results over similar techniques such as learning by predicting noise~\cite{bojanowski17} and learning by predicting color~\cite{larsson17}. The underlying idea is for the model to learn high-level semantic features in order to estimate the degree by which an image has been rotated. For instance, to predict if a chest CT is flipped horizontally, the model may rely on heart orientation among other image cues, or the model may learn to distinguish between the apex and base of the lung to predict if the chest CT is flipped vertically. However, predicting the degree of rotation makes sense as a surrogate task only if the underlying images follow a consistent geometry and have landmarks adequate for the task of rotation prediction. For example, similar to chest CTs, fundus images show a consistent geometry with distinct anatomical landmarks, such as the optic disc and macula, whereas a small image cube extracted from a CT scan may lack distinct landmarks to enable a reliable prediction of 3D rotation. For such applications, we resort to other surrogate supervision schemes such as patch reconstruction using a Wasserstein GAN~\cite{Arjovsky17} and image colorization using a conditional GAN~\cite{larsson17}. \figurename~\ref{fig:ss} illustrates the surrogate supervision schemes used in our applications.

\section{Applications}
\label{sec:apps}

To investigate the effectiveness of surrogate supervision, we considered 4 medical imaging applications consisting of classification and segmentation for both 2D and 3D medical images. Specifically, we studied False Positive Reduction (FPR) for nodule detection in chest CTs as a representative for 3D pattern classification, severity classification of diabetic retinopathy in fundus images as a representative for 2D pattern classification, lung lobe segmentation in chest CTs as a representative for 3D image segmentation, and skin segmentation in color tele-medicine images as a representative for 2D image segmentation. Table~\ref{tab:summary} summarizes the dataset for each application along with the selected surrogate supervision and experimental setup. Our \href{https://github.com/plantver0/isbi_2019_surrogate_sup_implementation_details/blob/master/index.md}{supplementary material} \cite{sm} provides further details on each studied application.

\begin{table*}
\centering
\begin{tabular}{lllll}
\multicolumn{4}{l}{\small \ding{61} denotes a fully labeled dataset; and thus, $D^u=Tr^l \cup V^l$}    & \textbf{Data split}  \\
\textbf{Task}&  \textbf{Dataset}& \textbf{Architecture}&  \textbf{Surrogate supervision} & $Tr^l|$$V^l|$$Te^l|$$D^u$  \\
\hline
FPR for nodule det.&  LIDC-IDRI \cite{armato11}&  8-layer 3D CNN \cite{ding17}& 3D patch reconstruction & 733$|$95$|$190$|$828{\ding{61}} CTs\\ 
Lung lobe seg.&  LIDC-IDRI \cite{armato11}&  Progressive dense V-Net \cite{Imran18}&  Rotation & 250$|$20$|$85$|$933 CTs\\
DR cls.&  Kaggle& Inception-resnet v2 \cite{SzegedyIV16}&  Rotation &70K$|$3K$|$15K$|$73K{\ding{61}} \\
Skin seg.& Private& resnet50-DeepLabV3+ \cite{deeplabv3plus2018} & Colorization & 1,435$|$100$|$163$|$330K \\
\hline
\end{tabular}
\caption{Summary of the tasks studied in this research. $D^u$ denotes the unlabeled dataset, whereas $Tr^l$, $V^l$, $Te^l$ denote the training, validation, and test subsets, respectively, of the labeled dataset. See Section~\ref{sec:apps} and \href{https://github.com/plantver0/isbi_2019_surrogate_sup_implementation_details/blob/master/index.md}{supplementary material} \cite{sm}.}
\label{tab:summary}
\end{table*}

Diabetic retinopathy (DR) classification requires assigning a severity level between 0 and 4 to each fundus image. For this, we have used inception-resnet v2~\cite{SzegedyIV16}. To pre-train this architecture, we used rotation as surrogate supervision by appending a fully connected layer with 3 neurons corresponding to $0^\circ$, $90^\circ$, and $270^\circ$ rotation followed by softmax with a cross entropy loss layer. Note that applying a $180^\circ$ rotation on the retina image from the left eye results in an image with appearance similar to the unrotated retina image from the right eye. In view of this ambiguity, we did not use a $180^\circ$ rotation during training with surrogate supervision. Nevertheless, rotation is a reasonable surrogate for DR because fundus images have a consistent geometry; thus, our model can learn the relative locations of structures such as the macula, optic disc, and retinal network by learning to predict the rotation.

The FPR task for nodule detection requires labeling each nodule candidate as either nodule or non-nodule. We use a 3D faster RCNN model to generate nodule candidates and further use the 3D CNN architecture suggested in~\cite{ding17} for FPR. The FPR model consists of 8 convolution layers with a pooling layer followed by dropout layer placed after every 2 convolution layers. FPR is commonly done based on local 3D patches around candidates, but these patches are often structurally sparse with no clear landmark; hence, they are unsuitable for the task of rotation estimation. Instead, a GAN~\cite{Arjovsky17} is utilized for the surrogate supervision task, where 3D patches are reconstructed using a generator network and we use the real-or-fake signal predicted by the discriminator network as the surrogate label. For this purpose, we use a U-Net-like network~\cite{RonnebergerFB15} as the generator and employ the above 8-layer architecture as the discriminator. Model training is conducted in a progressive manner as suggested by~\cite{abs-1710-10196} for generating higher quality 3D patches. Once converged, we use the discriminator as the pre-trained model for FPR. By learning to generate 3D patches, our model may learn about the continuity of vessels across slices as well as to distinguish nodules from vascular structures.

Lung lobe segmentation requires assigning every voxel in a chest CT to one of the 5 major lobes of the lung or background. For this, we use the progressive dense V-Net introduced in~\cite{Imran18}. Since thorax geometry is consistent in chest CT scans (assuming images are re-oriented to a common axis code), it makes sense to use rotation as surrogate supervision. However, since 3D rotation is computationally expensive, we resort to flipping along the $x$, $y$, and $z$ axes. The model is pre-trained by appending a global average pooling layer and a fully connected layer with 3 neurons where each neuron is followed by a sigmoid cross entropy loss layer. Essentially, each of the 3 neurons is responsible for predicting flipping along a particular axis. By learning to predict how the chest CT is flipped, the model may learn the global geometry of the lung and heart, which is likely to aid the target task of lung lobe segmentation. For instance, the model can learn to distinguish between the apex and base of the lung while learning to detect a flip along the $z$-axis, or it may learn the heart orientation while trying to detect a flip along the $x$-axis. 

Body skin segmentation is a multi-class segmentation problem for tele-medicine images, where each pixel is labeled as background, uncertain, face, hand, foot, limb, trunk, scalp, or anogenital. The tele-medicine images are taken by cell phone cameras and, thus, the body parts can appear in arbitrary orientations in the captured images, which naturally rules out the possibility of using orientation as surrogate supervision. Instead, we use image colorization for pre-training, by which the model can learn about the color and texture of the skin in an unsupervised fashion, which in turn can aid the target skin segmentation task. For image colorization, we use a conditional GAN~\cite{IsolaZZE16}, where the generator is a U-Net architecture based on resnet50-DeepLabV3+~\cite{deeplabv3plus2018} and the discriminator is a simple CNN model with 3 convolution layers. Once trained, the generator of the conditional GAN serves as a pre-trained model for skin segmentation.

\begin{figure}
\centering
\includegraphics[width=0.75\linewidth]{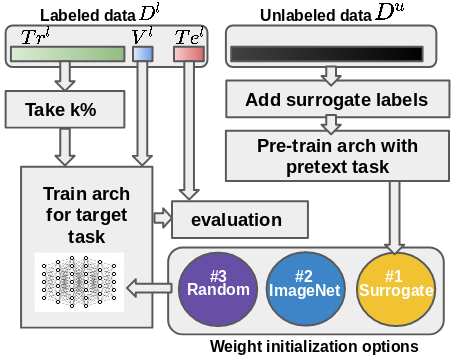}
\caption{The experimental protocol used in this study. \iffalse Note that the ImageNet model is available only for the 2D applications under study.\fi}
\label{fig:protocol}
\end{figure}

\section{Experiment Setup and Results}

\figurename~\ref{fig:protocol} illustrates the experiment setup in its general form for the studied applications. Specifically, for the 2D applications, we trained the models from scratch using Xavier's initialization, from an ImageNet pre-trained model, and from a model pre-trained by surrogate supervision. By doing so, we can compare the impact of surrogate supervision (done directly in the target domain) against transfer learning from a distant domain and training from scratch. For the 3D applications, we trained the models from scratch using Xavier's initialization and from a model pre-trained by surrogate supervision. Note that transfer learning from natural images is not possible for 3D medical imagining applications.
Also, to investigate the impact of surrogate supervision in the presence of limited labeled datasets, we have trained the models described above using $k$=10\%, 25\%, 50\%, and 100\% of the available labeled training data (see \figurename~\ref{fig:protocol}).

The data split can change from one application to another depending on whether the corresponding dataset is partially or fully labeled. If the entire dataset is labeled, we first divide it into 3 disjoint training, validation, and test subsets: $Tr^l$, $V^l$, and $Te^l$. We then form the unlabeled dataset by merging $Tr^l$ and $V^l$, followed by removing the labels of the target task, $D^u=Tr^l \cup V^l$. During training for the surrogate task, surrogate labels are assigned to unlabeled images. On the other hand, if the dataset is only partially labeled, we first divide the labeled part of the dataset into 3 disjoint training, validation, and test subsets. We then form the unlabeled dataset by merging the remaining unlabeled images, denoted $X$, with $Tr^l$ and $V^l$; i.e., $D^u=X \cup Tr^l \cup V^l$. As with the previous scenario, labels related to the target tasks are removed from the images placed in the unlabeled dataset. Table~\ref{tab:summary} specifies the size of each dataset split for each application under study.

\begin{figure}
\includegraphics[width=0.95\linewidth]{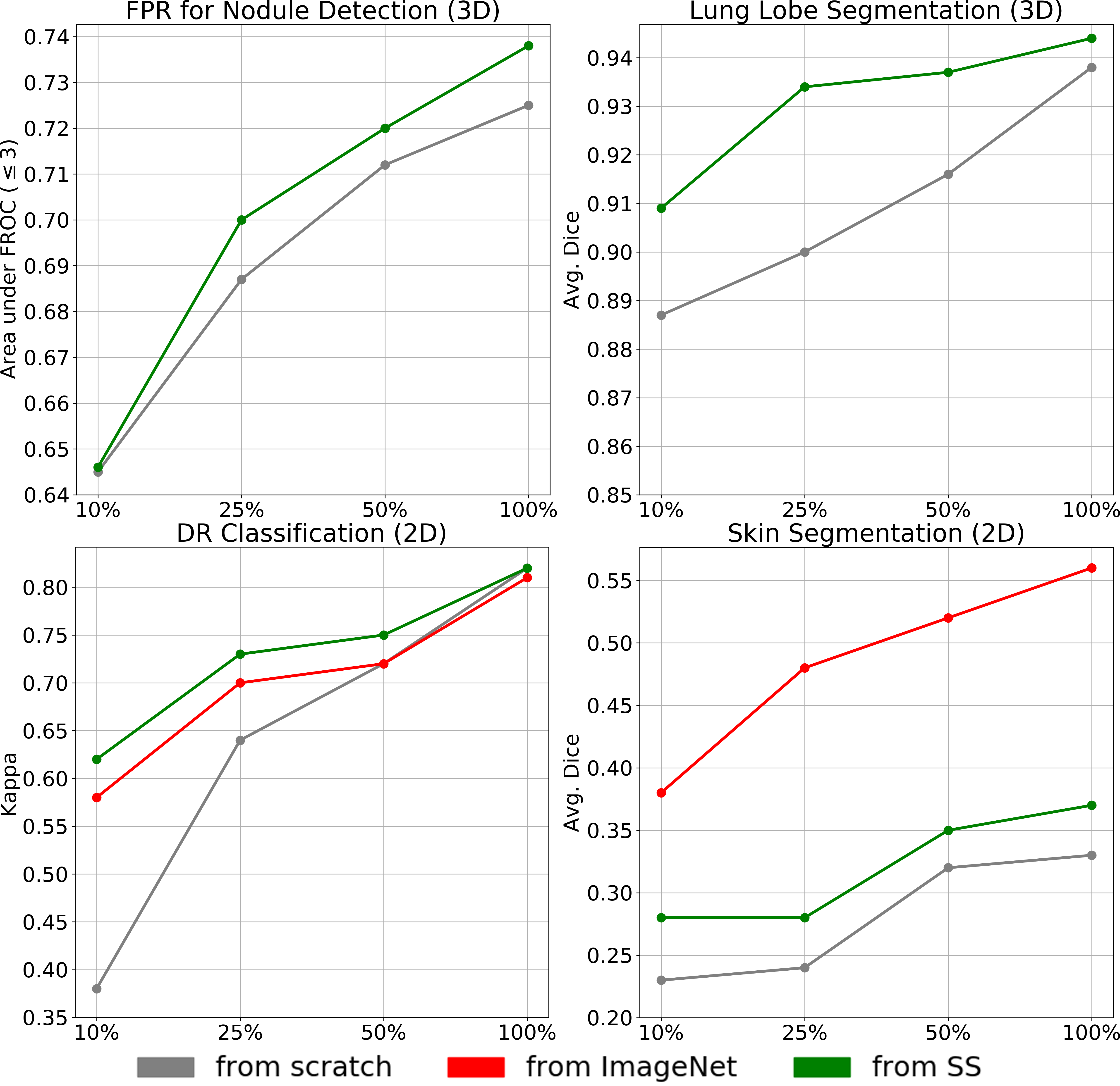}
\caption{Performance evaluation for the applications under study. \iffalse The horizontal axis shows the fraction of the training set used for model training.\fi Except for skin segmentation (see text), pre-training with surrogate supervision ({\bf SS}) is more effective than the ImageNet-trained model and random initialization.}
\label{fig:results}
\end{figure}

\figurename~\ref{fig:results} shows the results for the 4 applications under study. For DR classification, we used the Kappa statistic to measure the agreement between model predictions and ground truth, the average dice score for lung lobe segmentation and skin segmentation, and the area under the FROC curve (up to 3 false positives) for nodule detection. We see that pre-training with surrogate supervision enables the training of better-performing models, particularly when limited labeled data (10\% or 25\%) are used for training. However, such improvement tends to diminish \emph{in some applications} when the training set grows in size.

For 3D applications, pre-training with surrogate supervision shows marked improvement over training from scratch, particularly for lung lobe segmentation, which requires strong supervision and large quantities of labeled data. This is an important finding, because 3D architectures are commonly trained from scratch due to the scarcity of pre-trained 3D models, whereas they could have benefited from surrogate supervision. Our results suggest that pre-training 3D networks with surrogate supervision merits consideration as the first step towards solving 3D medical image analysis tasks.

Of the 2D applications under study, skin images are much closer to the domain of natural images than fundus images. This is because skin images show body parts often along with indoor and outdoor scenes in the background depending on camera-skin distance, whereas fundus images, which show only the retina, are quite distinct from natural images. The domain discrepancy between fundus images, skin images, and natural images is clearly reflected in our experimental results, inasmuch as pre-training with surrogate supervision is quite effective for the DR classification, but inferior to ImageNet weights for skin segmentation. It is noteworthy that, for both applications, pre-training with surrogate supervision is superior to random initialization. Our experiments suggest that pre-training with surrogate supervision in the target domain can be more effective than transferring weights from an unrelated domain even though the transferred weights are trained using much stronger supervision.

\vspace*{-3mm}
\section{Conclusion}

We investigated the effectiveness of surrogate supervision in training deep models for medical image analysis. Furthermore, we studied the impact of surrogate supervision with respect to the size of the training set. Our experimental results showed that models trained from weights pre-trained using surrogate supervision consistently outperformed the same models when trained from scratch. This is a key finding because 3D models in medical imaging have commonly been trained from scratch whereas they could have benefited from surrogate supervision. Our results further demonstrated that pre-training models in the medical domain was more effective than transfer learning from an unrelated domain (natural images). This finding highlights the practical value of unlabeled data in the medical imaging domain. We also observed that surrogate supervision was effective for the small training sets, but its impact tended to diminish for some applications as the sizes of the training sets grew. 

\bibliographystyle{IEEEbib}
\bibliography{refs}
\end{document}